\documentclass{arxivtmpl}

\def\1{\bm{1}}

\DeclareMathAlphabet{\mathsfit}{\encodingdefault}{\sfdefault}{m}{sl}
\SetMathAlphabet{\mathsfit}{bold}{\encodingdefault}{\sfdefault}{bx}{n}

\tcbuselibrary{skins,breakable}
\newtcolorbox{findingbox}{enhanced,breakable,frame hidden,colback=gred!6,borderline west={3pt}{0pt}{gred},sharp corners,boxsep=0pt,left=12pt,right=10pt,top=8pt,bottom=8pt,before skip=12pt,after skip=12pt}
\newtcolorbox{notebox}{enhanced,breakable,frame hidden,colback=gblue!6,borderline west={3pt}{0pt}{gblue},sharp corners,boxsep=0pt,left=12pt,right=10pt,top=8pt,bottom=8pt,before skip=12pt,after skip=12pt}
\newtcolorbox{tipbox}{enhanced,breakable,frame hidden,colback=ggreen!6,borderline west={3pt}{0pt}{ggreen},sharp corners,boxsep=0pt,left=12pt,right=10pt,top=8pt,bottom=8pt,before skip=12pt,after skip=12pt}
\newtcolorbox{warnbox}{enhanced,breakable,frame hidden,colback=gyellow!6,borderline west={3pt}{0pt}{gyellow},sharp corners,boxsep=0pt,left=12pt,right=10pt,top=8pt,bottom=8pt,before skip=12pt,after skip=12pt}

\tcbuselibrary{listings}
\newcommand{\boxfont}{\fontsize{9}{11}\selectfont}
\newtcblisting{promptbox}[1][]{%
  enhanced, listing only, breakable, frame hidden, width=\linewidth,
  colback=gblue!6, borderline west={3pt}{0pt}{gblue},
  sharp corners, boxsep=0pt,
  left=12pt, right=10pt, top=8pt, bottom=8pt, before skip=12pt, after skip=12pt,
  before upper={\def\promptlabel{#1}\ifx\promptlabel\empty\else{\sffamily\bfseries\color{gblue}#1}\par\smallskip\fi},   %
  listing options={basicstyle=\ttfamily\boxfont, escapeinside={(*}{*)}},
}

\newtcolorbox{transcriptbox}{%
  enhanced, breakable, frame hidden,
  colback=ggreen!6, borderline west={3pt}{0pt}{ggreen},
  sharp corners, boxsep=0pt, fontupper=\ttfamily\boxfont, halign upper=flush left,   %
  left=12pt, right=10pt, top=6pt, bottom=8pt, before skip=12pt, after skip=12pt,
}

\colorlet{codebg}{gyellow!6}

\hypersetup{
    colorlinks=true,
    linkcolor=linkcol,
    citecolor=linkcol,
    urlcolor=linkcol,
    pdfborderstyle={},
    pdfborder={0 0 0}
}
\renewcommand\Authfont{\sffamily\bfseries}
\renewcommand\Affilfont{\sffamily\mdseries}

\colorlet{perfhue}{gblue}
\colorlet{perf0}{white}
\colorlet{perf10}{perfhue!12}
\colorlet{perf20}{perfhue!22}
\colorlet{perf30}{perfhue!33}
\colorlet{perf40}{perfhue!44}
\colorlet{perf50}{perfhue!55}
\colorlet{perf60}{perfhue!66}
\colorlet{perf70}{perfhue!77}
\colorlet{perf80}{perfhue!88}

\newcommand{\RN}[1]{%
	\textup{\lowercase\expandafter{\it \romannumeral#1}}%
}

\SetKwInput{KwInput}{Input}
\SetKwInput{KwOutput}{Output}
\DontPrintSemicolon

\SetKwComment{Comment}{\color{ggreen}\# }{}

\SetKwProg{Function}{def}{:}{}

\SetKwProg{For}{for}{:}{}
\SetKwProg{If}{if}{:}{}

\usepackage{longtable}   %
\usepackage{xltabular}   %
\usepackage{fontawesome5}   %

\title{FrontierCS 2: A Benchmark for Computer Science R\&D}

\newcommand\fcsequal{%
  \fcsa{1}{*,$\dagger$}{Qiuyang~Mang}%
  \fcsa{2}{*,$\dagger$}{Wenhao~Chai}%
  \fcsa{1}{*}{Runyuan~He}%
  \fcsa{3}{*}{Dongruixuan~Li}%
  \fcsa{4}{*}{Zihan~Zheng}%
  \fcsa{5}{*}{Yujun~He}%
}
\newcommand\fcscontrib{%
  \fcsa{1}{}{Xuanyi~Zhou}%
  \fcsa{1}{}{Hao~Wang}%
  \fcsa{6}{}{Siddhant~Ray}%
  \fcsa{2}{}{Bo~Peng}%
  \fcsa{7}{}{Ruyi~Ji}%
  \fcsa{1}{}{Shuo~Yang}%
  \fcsa{1}{}{Haocheng~Xi}%
  \fcsa{1}{}{Boru~Chen}%
  \fcsa{1}{}{Xinyang~David~Han}%
  \fcsa{1}{}{Melissa~Pan}%
  \fcsa{1}{}{Mayank~Mishra}%
  \fcsa{8}{}{Boxi~Yu}%
  \fcsa{2}{}{Sijia~Liu}%
  \fcsa{1}{}{Kerem Akillioglu}%
  \fcsa{1}{}{Yi~Pan}%
  \fcsa{1}{}{Zirui~Wang}%
  \fcsa{1}{}{Mert Cemri}%
  \fcsa{1}{}{Diogo Antunes}%
  \fcsa{1}{}{Yichuan~Wang}%
  \fcsa{5}{}{Aiwei~Ivy~Zhang}%
  \fcsa{5}{}{Zhenyang~Chen}%
  \fcsa{11}{}{Qizheng~Zhang}%
  \fcsa{12}{}{Kaiyuan~Liu}%
  \fcsa{9}{}{Wenyuan~Huang}%
  \fcsa{11}{}{Sherry Ruan}%
  \fcsa{1}{}{Shu~Liu}%
  \fcsa{1}{}{Dacheng~Li}%
  \fcsa{2}{}{Zerui~Cheng}%
  \fcsa{2}{}{Zeyu~Shen}%
  \fcsa{10}{}{Shang~Zhou}%
  \fcsa{5}{}{Hanfei~Chen}%
  \fcsa{13}{}{Yuzhong~Zhang}%
  \fcsa{1}{}{Bohan~Lyu}%
  \fcsa{14}{}{Qu~Ao}%
  \fcsa{14}{}{Han~Zheng}%
  \fcsa{14}{}{Chonghe~Jiang}%
  \fcsa{15}{}{Jingzhuo~Hu}%
  \fcsa{18}{}{Xiaokun~Chen}%
  \fcsa{1}{}{Kunjun~Li}%
  \fcsa{16}{}{Zhiqing~Zhong}%
  \fcsa{16}{}{Haoye~Tian}%
  \fcsa{17}{}{Tong~Zhao}%
  \fcsa{17}{}{Yunzhong~He}%
  \fcsa{1}{$\dagger$}{Hanchen~Li}%
}
\newcommand\fcsadvisors{%
  \fcsa{}{}{Ofir~Press}%
  \fcsa{19}{}{Jiantao~Jiao}%
  \fcsa{11}{}{Kunle Olukotun}%
  \fcsa{2}{}{Karthik~R.~Narasimhan}%
  \fcsa{10}{}{Jingbo~Shang}%
  \fcsa{1}{}{Koushik~Sen}%
  \fcsa{1}{}{Matei~Zaharia}%
  \fcsa{1}{}{Alex~Dimakis}%
  \fcsa{1}{}{Ion~Stoica}%
  \fcsa{1}{}{Joseph~E.~Gonzalez}%
  \fcsa{1}{}{Alvin~Cheung}%
}
\newcommand\fcsaffiliations{%
  \fcsaffil{1}{UC Berkeley}%
  \fcsaffil{2}{Princeton University}%
  \fcsaffil{3}{University of Waterloo}%
  \fcsaffil{4}{New York University}%
  \fcsaffil{5}{Georgia Institute of Technology}%
  \fcsaffil{6}{University of Chicago}%
  \fcsaffil{7}{University of Michigan}%
  \fcsaffil{8}{University of Limerick}%
  \fcsaffil{9}{University of Wisconsin-Madison}%
  \fcsaffil{10}{UCSD}%
  \fcsaffil{11}{Stanford University}%
  \fcsaffil{12}{University of Washington}%
  \fcsaffil{13}{UNSW}%
  \fcsaffil{14}{MIT}%
  \fcsaffil{15}{University of Pennsylvania}%
  \fcsaffil{16}{Aalto University}%
  \fcsaffil{17}{Scale AI}%
  \fcsaffil{18}{Harbor Framework}%
  \fcsaffil{19}{NVIDIA}
}
\makeatletter
\newcommand\fcsaffil[2]{\@namedef{fcs@inst@#1}{#2}}
\fcsaffiliations
\newcount\fcs@n \newcount\fcs@cols \newdimen\fcs@wd
\newcommand\fcsa[3]{%
  \parbox[t]{\dimexpr\fcs@wd-6pt}{\raggedright
    {\fontsize{10}{12}\selectfont\strut#3\ifx\relax#2\relax\else\smash{\textsuperscript{\sffamily\color{ggrey}#2}}\fi\par}%
    {\fontsize{7.5}{9.5}\selectfont\color{ggrey}\strut\ifx\relax#1\relax\else\@nameuse{fcs@inst@#1}\fi\par}}%
  \hspace{6pt}\global\advance\fcs@n 1
  \ifnum\fcs@n=\fcs@cols \global\fcs@n=0 \par\vspace{5pt}\noindent\fi}
\newcommand\fcsmail[2]{%
  \parbox[t]{\dimexpr\fcs@wd-6pt}{\raggedright
    {\fontsize{10}{12}\selectfont\strut#1\par}%
    {\fontsize{8}{10}\selectfont\color{ggrey}\strut\href{mailto:#2}{\texttt{#2}}\par}}%
  \hspace{6pt}\global\advance\fcs@n 1
  \ifnum\fcs@n=\fcs@cols \global\fcs@n=0 \par\vspace{5pt}\noindent\fi}
\newcommand\fcscards[2]{\par{\fcs@cols=#1 \fcs@wd=\dimexpr\linewidth/#1\relax \global\fcs@n=0
  \setlength\parskip{0pt}\setlength\lineskip{0pt}\setlength\lineskiplimit{0pt}\noindent#2\par}}
\makeatother

\begin{document}

\noindent\includegraphics[width=110pt]{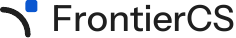}\par\vspace*{7pt}
\makeatletter\gdef\@author{{\Authfont FrontierCS Team}}\makeatother
\maketitle

\thispagestyle{firstpagestyle}

\begin{abstract}
{\sffamily\noindent
\begin{minipage}[t]{0.5\linewidth}{\small\bfseries\faGlobe\enspace Website}\par\vspace{1pt}{\small\href{https://frontier-cs.org}{frontier-cs.org}}\end{minipage}%
\begin{minipage}[t]{0.5\linewidth}{\small\bfseries\faGithub\enspace Code}\par\vspace{1pt}{\small\href{https://github.com/FrontierCS/FrontierCS-2}{github.com/FrontierCS/FrontierCS-2}}\end{minipage}\par}
\vspace{0.9\baselineskip}
{\sffamily\fontseries{eb}\normalsize\selectfont Abstract}\par\vspace{0.15\baselineskip}
\small
Can LLM agents make contributions to frontier computer science R\&D? We introduce FrontierCS 2, a benchmark of 48 open-ended tasks. We take a paper and the code its authors released and delete the paper's contribution. The deleted code becomes the human reference that sets the bar.
Earlier open-ended benchmarks give each agent a problem statement and grade one program after the run. FrontierCS 2 places the agent inside a research environment. The agent inherits a working repository and may submit a patch at any point. For each patch, a judge returns a development-suite score but keeps the evaluator, the held-out final suite and the reference private, and grades the final submission on that suite against the reference.
The tasks span 13 domains, from machine-learning systems and high-performance computing to databases, security and robotics. Beyond these 48 tasks, FrontierCS 2 keeps a private set of more than 100 tasks, to be announced in the future.
We evaluate 12 agents. The best one is Claude Opus 5.5, passing 58\% of its runs. No run passes on 16 tasks. A higher score is not the same as a pass: most runs that beat the reference on the combined final score still fall short on some metric or workload. On a shared capability scale for agents and tasks, the median task is roughly as difficult as what the strongest agent can handle.
\end{abstract}
\vspace{1.5em}

\begingroup\centering\captionsetup{type=figure}
\includegraphics[width=\linewidth]{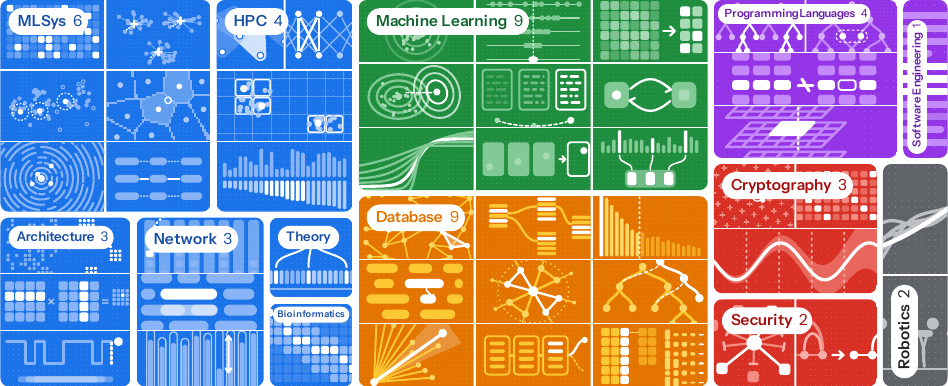}\par\vspace{0.4em}
\caption{\textbf{FrontierCS 2.} Each task removes a paper's key contribution; agents must beat the original code. \label{fig:overview}}\par\endgroup
\write-1{TEASER-FIT page=\thepage}   %

\clearpage
\section*{Author list}
{\setlength{\parindent}{0pt}\raggedright\sffamily
\vspace{0.5em}\fcscards{4}{\fcsequal}\vspace{1em}
\fcscards{4}{\fcscontrib}\vspace{0.6em}
{\fontsize{8}{10}\selectfont\color{ggrey}\textls[80]{ADVISORS}\par}\vspace{0.5em}
\fcscards{4}{\fcsadvisors}\vspace{0.8em}
{\fontsize{8}{10}\selectfont\color{ggrey}\textls[80]{CORRESPONDENCE}\par}\vspace{0.5em}
{\small\hypersetup{urlcolor=ggrey}$^\dagger$\,\href{mailto:qmang@berkeley.edu}{qmang@berkeley.edu}\quad \href{mailto:wenhao.chai@princeton.edu}{wenhao.chai@princeton.edu}\quad \href{mailto:lihanc@berkeley.edu}{lihanc@berkeley.edu}\par}}
\clearpage   %
\section{Introduction}
\label{sec:intro}

\begin{figure}[t]
\centering
\includegraphics[width=\linewidth]{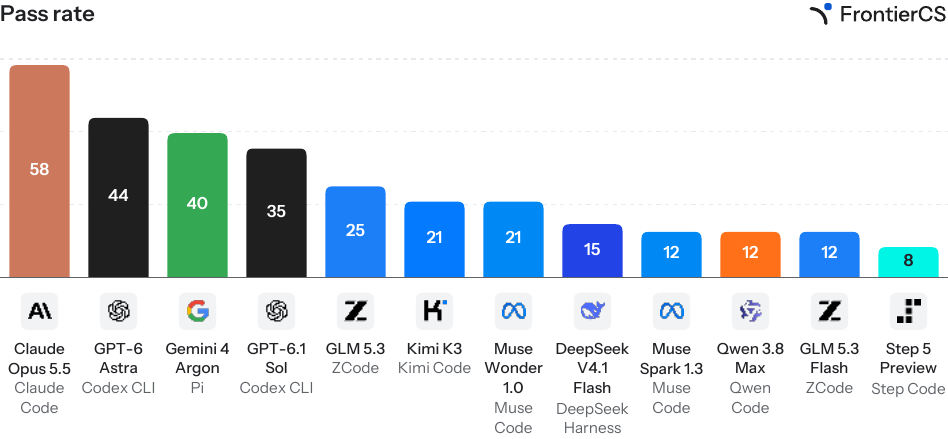}
\caption{\textbf{Pass rate.} Share of each agent's runs whose final submission beats the human reference on every metric and held-out workload; the harness is in gray.}
\label{fig:pass-rate}
\end{figure}

Consider a researcher who picks up a published system to make it better. They inherit a working codebase, change one component, test the change on the workloads the paper reported, and are judged against the paper's own implementation. Much of research and development in computer science looks like this. We ask whether an LLM agent can do the same.

Existing open-ended benchmarks answer a narrower question. On FrontierCS~\citep{mang2026frontiercs}, ALE-Bench~\citep{imajuku2025alebench} and AtCoder heuristic contests~\citep{liu2026slowdown}, agents stay below the top human experts. These benchmarks share one design: the agent receives a problem statement and writes one program, and an evaluator scores it once, after the run ends.

Benchmarks that do place the agent in a real environment are narrow in another way. SWE-bench~\citep{jimenez2024swebench} and MLE-bench~\citep{chan2025mlebench} grade against tests or a leaderboard. ResearchGym~\citep{garikaparthi2026researchgym}, MLS-Bench~\citep{lyu2026mlsbench} and PaperBench~\citep{starace2025paperbench} stay within machine learning, and those that span other fields grade against a rubric or a reimplementation rather than the authors' own code.

\begin{table}[t]
\caption{\textbf{FrontierCS and FrontierCS 2.} Each task moves into a paper's own repository, with the authors' code as the bar.}
\label{tab:fcs-versions}
\centering\small
\begin{tabular*}{\linewidth}{@{\extracolsep{\fill}}lll@{}}
\toprule
 & FrontierCS & FrontierCS 2 \\
\midrule
Task & A problem statement & A repository with a paper's contribution deleted \\
Output & One program & A patch \\
Bar & No known optimum & The authors' code, on held-out workloads \\
Feedback & None & A development score for every patch \\
Domains & Algorithms and six research areas & 13 computer-science domains \\
Metric & Mean score & Pass rate, Mean score, and FCI \\
\bottomrule
\end{tabular*}
\end{table}

In this paper, we evaluate agents against the published code itself, inside the repository it came from, across computer science. FrontierCS 2 is the second version of FrontierCS (\Cref{tab:fcs-versions}). We take the code a paper's authors released, delete the paper's contribution, and ask the agent to rebuild it so that it beats the original (\Cref{fig:overview}). We call the deleted code the human reference and measure it on the paper's own workloads (\Cref{sec:method}). The agent works as a researcher would. It edits the code, submits a patch whenever it likes, and gets back a score on a development suite from a judge. The evaluator, the held-out final suite and the human reference stay with the judge (\Cref{sec:scoring}). At the end, the agent names one patch as final, and the judge grades that patch on the held-out suite. A run passes only when the patch beats the human reference on every metric and every held-out workload.

We evaluate 12 models, each paired with one agent harness, with one run per task. The best agent, Claude Opus 5.5 in Claude Code, passes 58\% of its runs, and every other agent passes fewer than half of its runs (\Cref{fig:pass-rate}). No agent passes any of 16 tasks. To compare agents and tasks on one scale, we fit the FrontierCS Capabilities Index (FCI) together with a matching difficulty index. The median task sits at about the strongest agent's FCI (\Cref{sec:exp-eci}). The leading agent also changes with the cost budget per run (\Cref{sec:scaling}).

In summary, our contributions are:
\begin{itemize}[topsep=2pt,itemsep=1pt]
\item Tasks built by subtraction: we delete a paper's contribution from its repository and keep the deleted code as the bar (\Cref{sec:method}).
\item A judge that scores patches during a run and keeps the evaluator, the held-out suite and the reference private (\Cref{sec:scoring}).
\item 48 tasks in 13 domains, with 12 agents evaluated (\Cref{sec:experiments}) and three case studies (\Cref{sec:cases}), and a private set of more than 100 tasks, to be announced in the future.
\end{itemize}

\clearpage
\setcounter{tocdepth}{2}
\begingroup
\hypersetup{linkcolor=black}
\tableofcontents
\endgroup
\clearpage

\section{Related work}
\label{sec:related-work}

\paragraph{Open-ended coding benchmarks.}
Unlike competition benchmarks with known solutions, such as LiveCodeBench Pro~\citep{zheng2025livecodebenchpro}, benchmarks that score a program on an objective with no known optimum now span heuristic contests~\citep{imajuku2025alebench}, combinatorial optimization~\citep{chen2026heurigym,li2025npengine}, GPU kernels~\citep{ouyang2025kernelbench}, and algorithmic and research problems~\citep{mang2026frontiercs}. Speed-up benchmarks start from a codebase and measure against a fixed baseline: the human nanoGPT speedrun record~\citep{zhao2025speedrunning,bakouch2026measuring} or a naive PyTorch inference server~\citep{yeon2026inferencebench}. FrontierCS 2 instead grades most of its tasks against the human reference, code the paper's authors released, and scores each patch during the run.

\paragraph{Agent benchmarks for research and engineering.}
A second family scores what an agent produces inside an environment. SWE-bench~\citep{jimenez2024swebench} and Terminal-Bench~\citep{merrill2026terminalbench} grade with held-out tests, MLE-bench~\citep{chan2025mlebench} with Kaggle leaderboards, and RE-Bench~\citep{wijk2025rebench} against human experts given the same time budget. PostTrainBench~\citep{rank2026posttrainbench} measures the downstream score of a post-trained model, and DeployBench~\citep{wang2026deploybench} checks whether a research artifact runs its designated experiment. EdgeBench~\citep{zhu2026edgebench} also gives the agent feedback during a run. PaperBench~\citep{starace2025paperbench} grades replication from scratch with a rubric, while AIRS-Bench~\citep{lupidi2026airsbench} and ResearchGym~\citep{garikaparthi2026researchgym} compare against published machine-learning results. RSI-Forge~\citep{wang2026rsiforge} turns papers from many fields into tasks whose baseline is an agent's reimplementation of each paper's method. Into the Unknown~\citep{kim2026intounknown} grades follow-up papers on unpublished directions against expert criteria, and ReuF2F~\citep{zheng2026reuf2f} poses formalized undergraduate research problems. None of these benchmarks sets its bar by measuring the original authors' code. MLS-Bench~\citep{lyu2026mlsbench} is the closest in design: each of its 140 machine-learning tasks across 12 areas asks the agent to improve one named component. We extend that design beyond machine learning to 13 computer-science domains, from databases to cryptography.

\paragraph{Building tasks from existing artifacts.}
Methods that build tasks at scale start from an artifact that already works and break it in a controlled way. SWE-bench pairs a pull request with the tests it fixed, and SWE-smith~\citep{yang2025swesmith} injects bugs into a repository so that existing tests fail. FrontierSmith~\citep{he2026frontiersmith} mutates closed competitive-programming problems into open-ended ones, and BenchEvolver~\citep{wu2026benchevolver} evolves solved problems into harder variants. DiscoGen~\citep{goldie2026discogen} instead generates algorithm-discovery tasks procedurally. Like ResearchGym, we remove a paper's method from its repository. Unlike it, our oracle is the removed code itself, and our tasks span 13 domains.

\clearpage   %
\section{Task building}
\label{sec:method}

\begin{figure}[t]
\centering
\includegraphics[width=\linewidth]{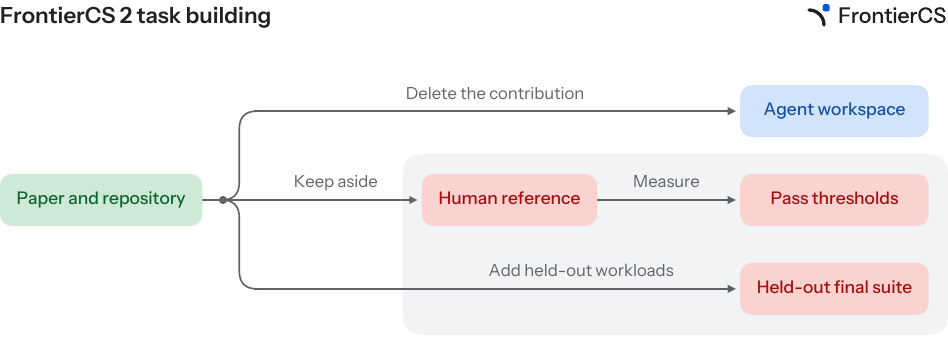}
\caption{\textbf{FrontierCS 2 task building.} We delete the paper's contribution from its own repository and give the agent what remains. The judge keeps the deleted code as the human reference, the pass thresholds measured from it on the paper's workloads, and a final suite with held-out workloads.}
\label{fig:construction}
\end{figure}

We build the tasks by subtraction from the code the paper's authors released (\Cref{fig:construction,tab:sources}).\footnote{Except Structured LWE, an original cryptanalysis challenge (\Cref{app:sources}).} We locate the code that implements the paper's contribution, delete it, and leave everything it depends on unchanged. Take SVG-EAR~\citep{zhou2026svgear}, which we build from the Sparse VideoGen repository\footnote{\url{https://github.com/svg-project/Sparse-VideoGen}}. We delete the error-aware attention class and three of its routines. These routines average each cluster, choose the cluster pairs to compute, and compute those pairs while estimating the rest. The agent receives this repository with the deleted functions still declared and called, and it must write them. It may edit only the files that held the contribution. The human reference is a private patch that restores the deleted code. Subtraction gives each task built this way an oracle that already runs in its repository on the paper's own workloads.

We set each metric's pass threshold from the reference's own measurements on the workloads the paper reported. For SVG-EAR, latency must be at most 0.96 times the reference's latency, PSNR must stay within 0.25~dB of the reference value, SSIM within 0.01 and LPIPS within 0.005. The judge scores every submission on a development suite of these workloads. Only the final submission is scored on a final suite, which adds held-out instances the agent never sees (different seeds, queries or input graphs).

\clearpage
\section{Judge}
\label{sec:scoring}

\begin{figure}[t]
\centering
\includegraphics[width=\linewidth]{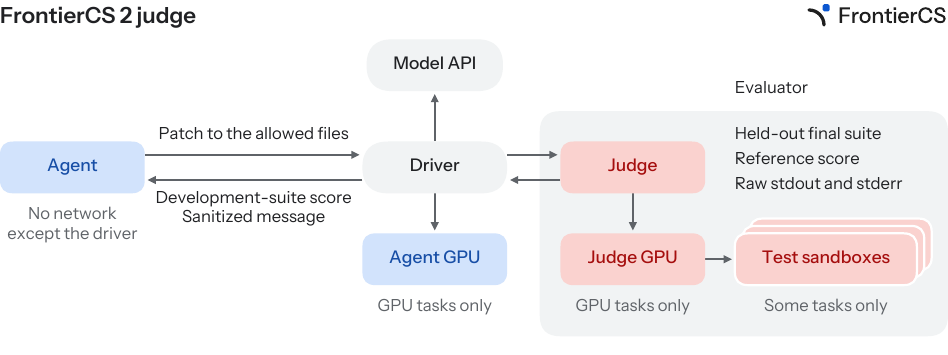}
\caption{\textbf{FrontierCS 2 judge.} Each colored box is a sandbox. The agent's sandbox reaches only the driver. The agent sees only a development-suite score and a sanitized message per patch; all that decides its result stays with the judge.}
\label{fig:evaluation}
\end{figure}

The agent and the judge hold separate information (\Cref{fig:evaluation}). The agent holds the repository; the judge holds the evaluator, the held-out final suite and the human reference. The agent has five hours and may submit a patch to the allowed files at any time. The judge returns the patch's score on the development suite and a message that counts the development workloads meeting each metric's threshold, without the reference score or the evaluator's raw output.

They also run apart, each in its own sandbox, linked only by a driver started for the run. The agent's sandbox reaches no network but the driver, which forwards its model calls and patches, so it cannot download the authors' code. On GPU tasks, the agent may start a development GPU through the driver, while only the judge uses the GPU that measures the patch against the human reference. Some tasks also give the judge test sandboxes: on Continuum~\citep{li2025continuum}, a vLLM serving task, they run SWE-bench episodes against the patched vLLM.

Development-suite scores guide the agent and do not count toward its result. At the end of the run, the agent names one submission as final, and the judge evaluates that patch alone on the held-out suite. We do not pick its best development-suite submission for it, since choosing the patch is part of a researcher's work. We say a submission beats the reference on a metric when it meets that metric's threshold. A run passes when its final submission beats the reference on every metric across every held-out workload. The judge also assigns a final score from 0 to 100 on a scale each task defines.\footnote{Two examples. KV Compaction scores each of its 24 articles as the share of questions answered correctly times $0.75 + 0.25e$, where $e = \min(1, \max(0, 0.5 + \log_{10} r / 6))$ and $r$ is the agent's speedup over the human reference, discounted when its compacted cache is larger (the task's bytes mode swaps the roles of time and size). Progent scores each of its four suites as the utility under attack times the share of attacked episodes that are not breached. Both report 100 times the mean.} A run without a saved final submission scores 0.

\clearpage

\section{Evaluation results}
\label{sec:experiments}

\subsection{Experimental setup}
\label{sec:exp-setup}

\begin{figure}[t]
\centering
\includegraphics[width=\linewidth]{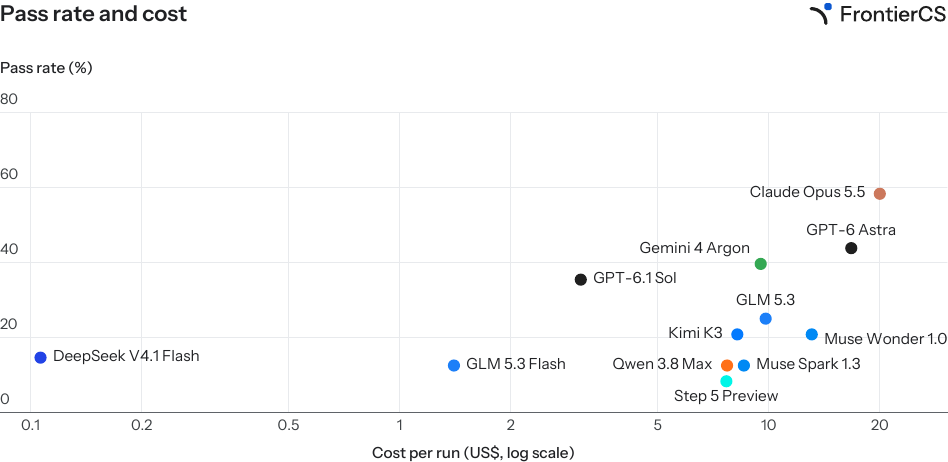}
\caption{\textbf{Pass rate and cost.} Each point is one agent: pass rate on the y-axis against mean cost per run in US\$ on a log-scale x-axis. GPT-6.1 Sol passes less often than GPT-6 Astra and sits far to its left on the cost axis.}
\label{fig:pass-cost}
\end{figure}

\paragraph{Agents.} An agent pairs a model with a harness, the program that runs the model's tool calls and submits its patches to the judge. We evaluate 12 models, each in one harness. We name each agent by its model, but the harness is part of what we measure. \Cref{tab:agents} lists each agent's harness, version, reasoning effort and output limit.

\paragraph{Tasks and runs.} A run is one agent on one task, from the starting repository to its final submission. Each of the 12 reported agents runs each task once. The provider refused two Gemini 4 Argon runs before a final submission. Like every run without a passing final submission, they count as failures, in the pass rate and in FCI alike.

\begin{figure}[t]
\centering
\includegraphics[width=\linewidth]{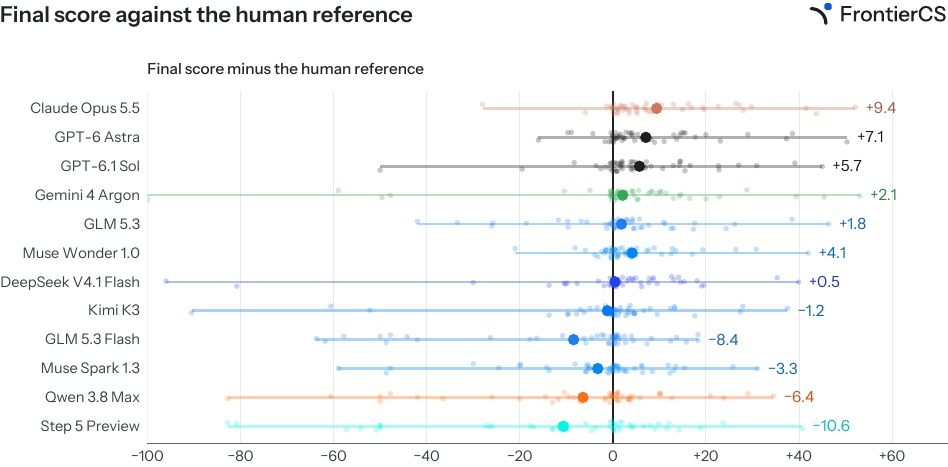}
\caption{\textbf{Final score against the human reference.} Each faint point is one run's final score minus the human reference's on the same task, one row per agent; the line spans each agent's range and the dot marks its mean. The vertical line at 0 marks a tie.}
\label{fig:score-ref}
\end{figure}

\paragraph{Metrics.} Our main metric is the pass rate, the share of an agent's runs that pass. We also report an agent's mean score, its mean margin over the human reference. This is the average over its runs of the final score minus the reference's score on the same task (0 is a tie). \Cref{fig:pass-cost} sets each agent's pass rate against its mean cost per run. A higher final score does not mean a pass, and a pass does not need a higher score (\Cref{fig:score-ref}). Most runs whose final score exceeds the human reference's miss a threshold on at least one metric or workload. Conversely, some passing runs score at or below the reference's final score, since the final score is one number on each task's own scale, while a pass needs every metric to hold at once.

\subsection{FrontierCS Capabilities Index (FCI)}
\label{sec:exp-eci}

The FrontierCS Capabilities Index (FCI) puts the agents on one capability scale, and the FrontierCS Difficulty Index (FDI) puts the tasks on the same scale. FCI adapts Epoch AI's Epoch Capabilities Index~\citep{ho2025rosetta}, an item response model. We fit it to per-metric outcomes and to score margins over the human reference, rather than to raw scores.

One joint fit takes two observations from every run: which of the task's metrics the run beats the human reference on, and the margin of its final score over the reference's. The metric outcomes tell the fit how hard each part of a task is, and the margin tells it how close a failed run came. Both share one fitted run effect, which says how far that run lands above or below its agent's strength.

Each agent has one strength, which we report as its FCI: ten points per logit, with Kimi K2.7 at 0. Kimi K2.7, run in Kimi Code, is a thirteenth agent that serves only as this 0 point: its runs enter the fit, but we do not report it. A task's FDI is the FCI at which an agent passes it half the time. The human reference has no FCI, since its code cannot beat itself. Intervals are 90\% bootstrap intervals over the tasks, and \Cref{app:fci} gives each step of the fit.

\begin{figure}[t]
\centering
\includegraphics[width=\linewidth]{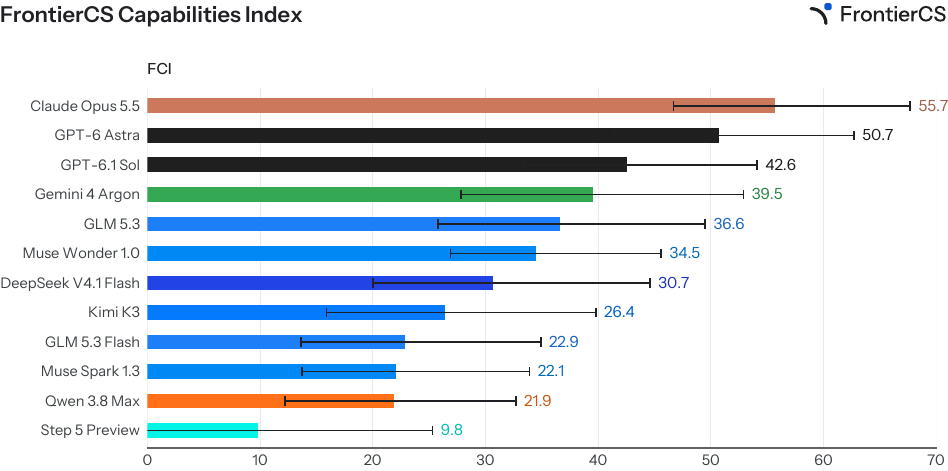}
\caption{\textbf{FrontierCS Capabilities Index.} Each bar shows an agent's FCI from 0 with its 90\% bootstrap interval as a whisker. The intervals separate only agents that sit far apart on the capability scale.}
\label{fig:fci}
\end{figure}

Claude Opus 5.5 leads FCI, followed by GPT-6 Astra and GPT-6.1 Sol, and Step 5 Preview is last (\Cref{fig:fci}); FCI and the pass rate order the agents almost alike. The tasks span a wider range of difficulty than the agents do: the median task sits at about the strongest agent's FCI. Each task's FDI is uncertain, however, so we do not order individual tasks by it. FDI measures how hard a task is to pass, and across tasks it follows the pass rate closely.

\subsection{Test-time scaling}
\label{sec:scaling}

\label{sec:exp-rules}Test-time scaling asks what FCI an agent would reach under a cost cap per run. At budget $B$, a run that finished within $B$ keeps its held-out result, and a run still going at $B$ counts as one without a final submission, which scores 0. With the task parameters fixed at their FCI fit (\Cref{sec:exp-eci}), the FCI at $B$ is the strength that best fits these outcomes (\Cref{app:fci}). A run's outcome can only improve as $B$ grows, so no curve falls, and each ends at the agent's FCI.

\begin{figure}[t]
\centering
\includegraphics[width=\linewidth]{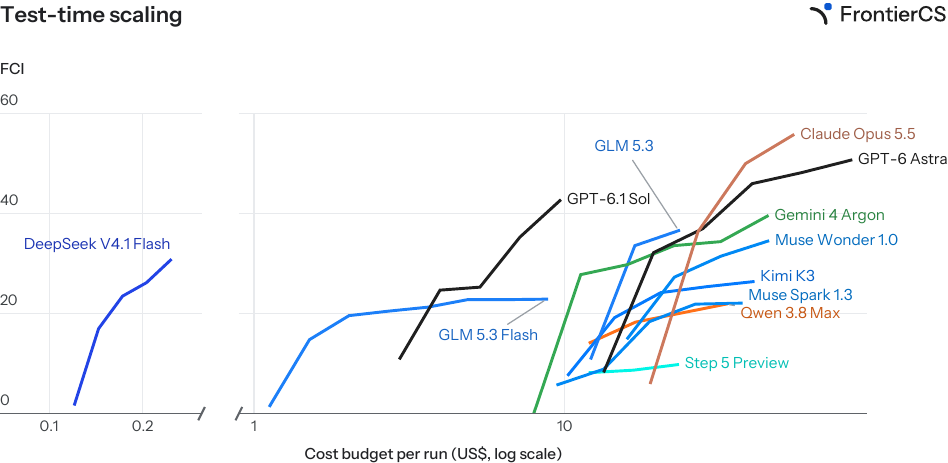}
\caption{\textbf{Test-time scaling.} FCI of each agent against the cost budget per run, in US\$ on a log scale; a run still going at a budget counts as one without a final submission. Each curve starts at the first budget at which its FCI is at least 0 and ends at the agent's FCI. Two slanted strokes mark an empty budget range skipped on the axis. The leader changes with the budget.}
\label{fig:scaling}
\end{figure}

\label{sec:exp-scaling}Which agent leads depends on the budget (\Cref{fig:scaling}). At the smallest budgets, DeepSeek V4.1 Flash leads. Its costs are lower bounds, so this lead depends on how we price its tokens. Claude Opus 5.5 leads at the largest budgets.

\section{Case studies}
\label{sec:cases}
\subsection{Two checksums shrink private lookups}
\label{sec:case}

\begin{figure}[t]
\centering
\includegraphics[width=\linewidth]{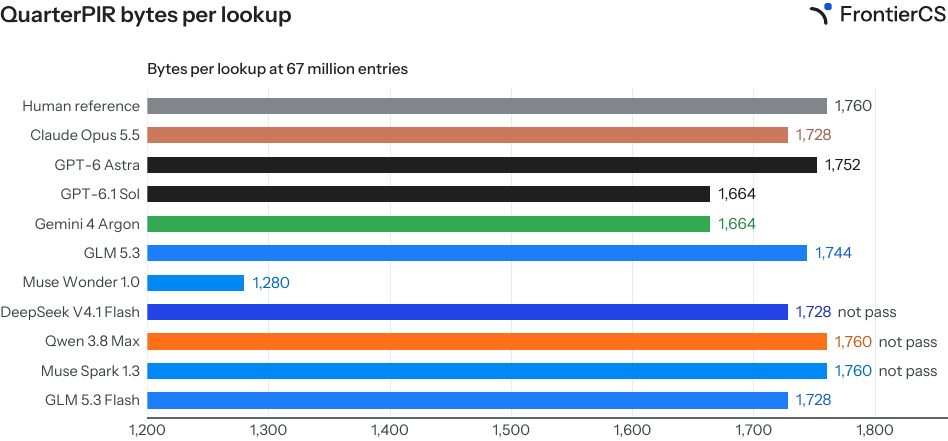}
\caption{\textbf{QuarterPIR bytes per lookup.} Bytes per private lookup on the 67-million-entry database; the x-axis starts at 1,200. Muse Wonder 1.0's checksums remove far more of the reference's bytes than any other passing run does. Kimi K3, far below the rest and not passing, is left out. Step 5 Preview's run timed out.}
\label{fig:case-quarterpir}
\end{figure}

Private information retrieval lets a client read one entry of a database held by a server, without the server learning which entry it read. QuarterPIR~\citep{ghoshal2024quarterpir} makes this cheap. The client first builds a compact summary of the database; afterwards, each lookup costs one short request and one short reply. In our task the agent rewrites how the server answers requests, and it may also change the messages and the client. A run passes if every lookup returns the right entry and the run beats the human reference on the bytes each lookup sends, on every timing, and on the memory the client keeps.

In the human reference, every reply carries values the client rarely needs. Along with the answer, the server returns one parity value for each group of the database. The client needs only one of them, to correct the hint it refreshed after its previous lookup. Asking for that one alone, however, would tell the server where its previous lookup went. Muse Wonder 1.0 replaces these values with two checksums, the construction that lets a RAID-6 disk array rebuild two failed disks. Since the client already knows the only two groups in which its own values can differ from the server's, two checksums are enough to recover the one it needs. This shrinks every reply, and therefore every lookup. Privacy is unchanged, because the task lets the server see only a public function of the request the authors' client would send.

Muse Wonder 1.0's run passes with a final score above the human reference's. Most agents pass this task, but no other passing run shrinks the lookup by nearly as much (\Cref{fig:case-quarterpir}). We found this idea only by reading the patches, since neither of our measures picks it out. A pass counts the same whether a run saves a few bytes or the most. GPT-6.1 Sol even scores slightly higher than Muse Wonder 1.0 with a far smaller saving.

\subsection{Two faster clocks for a packet scheduler}
\label{sec:case-bbq}

\begin{figure}[t]
\centering
\includegraphics[width=\linewidth]{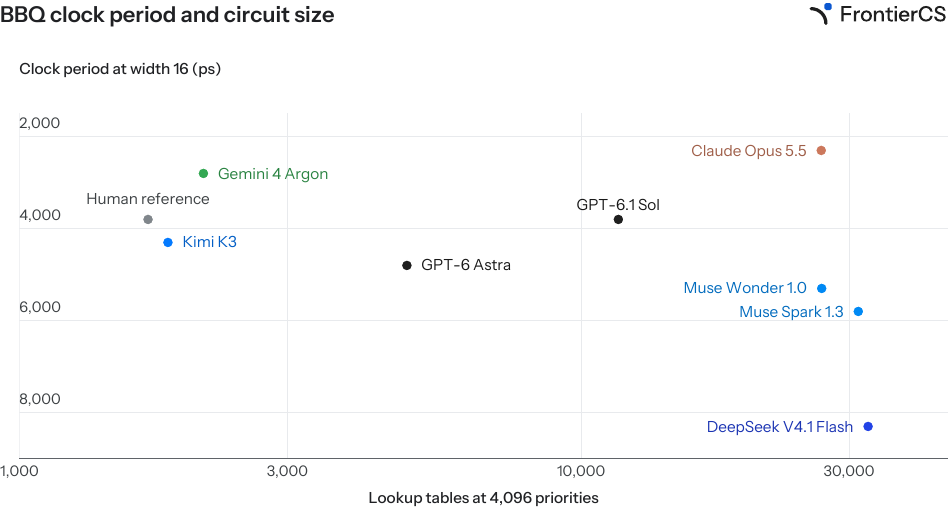}
\caption{\textbf{BBQ clock period and circuit size.} Each point is one agent's circuit on the largest tree, 4,096 priorities. The x-axis gives its size in lookup tables and the y-axis its clock period, with shorter, faster clocks at the top; the grey dot marks the human reference. Claude Opus 5.5's circuit has the shortest clock period and is among the largest. Agents without a correct circuit at this size are left out.}
\label{fig:case-bbq}
\end{figure}

A network switch or network card has to decide, for every packet it sends, which waiting packet goes next. Programmable schedulers express this decision as a priority queue: each packet carries a priority, and the queue must hand out the most urgent one millions of times per second. BBQ~\citep{atre2024bbq} builds such a queue as an FPGA circuit. It marks the occupied priorities in a tree of bitmaps. Each bit keeps a count of the packets below it, so that the bit can be cleared when its last packet leaves. In our task the agent rewrites the logic of the tree's levels. A run passes if the circuit is correct and beats the human reference on throughput and on worst-case latency at every tree size, including a larger one that the agent never sees during development.

On the smaller trees, the slowest step of each clock cycle in the human reference begins with a counter read from on-chip memory. The operation just ahead may have changed the same counter a moment earlier. The circuit therefore has to check the value it read against recent updates before it can use it. Gemini 4 Argon keeps the memories but settles that check while the read is still in flight, leaving little work for when the value arrives. Claude Opus 5.5 takes the other route. It moves every counter from memory into registers and updates a whole level of the tree in a single cycle, so that on the development trees no operation waits for a read.

Only these two and GPT-6.1 Sol pass, and the two designs trade speed for size (\Cref{fig:case-bbq}). Both handle more operations per second than the human reference on every tree. On the largest tree, however, the circuit of Claude Opus 5.5 uses the most lookup tables of the passing circuits. Its design also does not serve trees as large as those BBQ was designed for. Gemini 4 Argon keeps roughly the human reference's circuit size and memory layout. Registers alone do not explain Claude Opus 5.5's clock. Other agents also move the counters into registers, and on the largest tree they end up slower than the human reference. The final score ranks Claude Opus 5.5 first, because it rewards speed and ignores circuit size.

\subsection{A measured budget for sparse video attention}
\label{sec:case-svgear}

\begin{figure}[t]
\centering
\includegraphics[width=\linewidth]{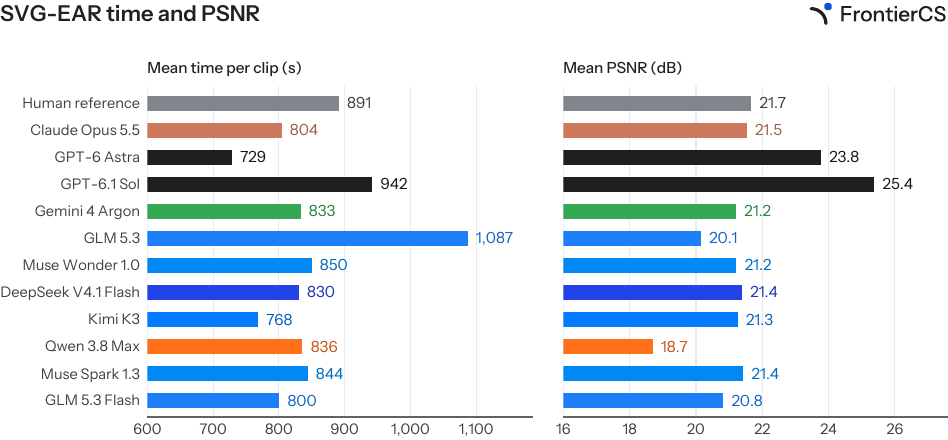}
\caption{\textbf{SVG-EAR time and PSNR.} Each run's mean time per clip and mean PSNR against full attention on the held-out clips, with the human reference's in grey; the x-axes start at 600 s and 16 dB. Only GPT-6 Astra is both faster and closer to full attention than the human reference on average, and it still fails. Step 5 Preview, far slower, is off the chart.}
\label{fig:case-svgear}
\end{figure}

Video diffusion models spend most of their time in attention. SVG-EAR~\citep{zhou2026svgear}, our running example in \Cref{sec:method}, computes some blocks of attention exactly and estimates the rest from cluster centroids. Within a fixed budget, it computes the blocks whose estimates would err the most. A run passes only if every held-out clip is faster than with the human reference and stays within narrow bands of its PSNR, SSIM and LPIPS.

No run passes, and GPT-6 Astra comes closest. It keeps the authors' design but measures the error instead of predicting it. Every few denoising steps, it computes the exact attention of a few real queries and rebuilds their output under several budgets. Each attention head then runs with the smallest budget whose error stays under a bound. Astra also corrects a bias of the centroid estimate that the paper leaves in. Its videos come out faster than the human reference's on every clip and, on average, closer to full attention (\Cref{fig:case-svgear}). It still fails on image quality in two clips. Speed was not the obstacle on this task. Most runs beat the reference's time on every clip, and no run stays within every image band.
\section{Conclusion}
\label{sec:conclusion}

We built FrontierCS 2 to ask whether LLM agents can make contributions to frontier computer science R\&D. Each task measures an agent against the code that researchers released with their paper, in that code's own repository. Beyond these 48 tasks, FrontierCS 2 keeps a private set of more than 100 tasks, to be announced in the future.

The case studies show what the pass and the final score capture and what they miss. On SVG-EAR, a run faster than the human reference on every clip and closer to full attention on average still fails on image quality in two clips (\Cref{sec:case-svgear}). On QuarterPIR, neither measure separates the run that removes far more bytes per lookup from the other passing runs (\Cref{sec:case}). On BBQ, the final score ranks the largest passing circuit first, because it ignores circuit size (\Cref{sec:case-bbq}). We think benchmarks like this one will need to report such side costs next to the pass.

FCI and FDI are the measures we recommend for benchmarks like this one. They place agents and tasks on one scale, use every metric outcome and score margin of a run rather than one number, and come with intervals that show which differences the data support (\Cref{sec:exp-eci}).

\newpage
\section*{Acknowledgments}
We give special thanks to Modal for its strong support of this work, including the compute resources it sponsored.

\vspace{3pt}\noindent\includegraphics[height=1.4em]{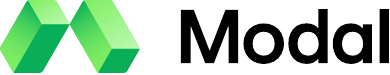}

\vspace{3pt}\noindent We also thank the Laude Institute's Slingshot program and Scale AI for compute credits and community support. We thank Kaggle for sponsoring FrontierCS.

\clearpage

\bibliography{references}

\clearpage
\appendix
\crefalias{section}{appendix}   %
\section*{Appendix}   %
\section{Task sources}
\label{app:sources}

\Cref{tab:sources} lists the paper behind each of the 48 tasks. The three FlashLib tasks share one source, and so do the two FARGO tasks. Structured LWE is the one task not built from a paper's code: it is an original cryptanalysis challenge in which the agent attacks 200 published instances, and any secret that meets the published conditions counts. Since these instances are public, its final suite has no held-out part.

\begingroup\footnotesize\setlength{\LTcapwidth}{\linewidth}
\begin{longtable}{@{}l@{\hspace{8pt}}l@{\hspace{8pt}}>{\raggedright\arraybackslash}p{0.49\linewidth}@{}}
\caption{\textbf{Task sources.} The paper each task is built from.}\label{tab:sources}\\
\toprule
Domain & Task & Paper \\
\midrule
\endfirsthead
\toprule
Domain & Task & Paper \\
\midrule
\endhead
\bottomrule
\endfoot
Database & ScaleGPM & Accurate and Fast Approximate Graph Pattern Mining at Scale~\citep{arpacidusseau2024scalegpm} \\
 & Yannakakis & Instance-Optimal Acyclic Join Processing Without Regret: Engineering the Yannakakis Algorithm in Column Stores~\citep{bekkers2025yannakakis} \\
 & FARGO Query Stopping & FARGO: Fast Maximum Inner Product Search via Global Multi-Probing~\citep{zhao2023fargo} \\
 & Spectrum & Spectrum: Speedy and Strictly-Deterministic Smart Contract Transactions for Blockchain Ledgers~\citep{chen2024spectrum} \\
 & SCOPE & Fast Local Subgraph Counting~\citep{li2024scope} \\
 & Persistent Treap & Efficient Concurrent Updates to Persistent Randomized Binary Search Trees~\citep{hou2025treap} \\
 & FARGO Query Core & FARGO: Fast Maximum Inner Product Search via Global Multi-Probing~\citep{zhao2023fargo} \\
 & Seer & Seer: Accelerating Blockchain Transaction Execution by Fine-Grained Branch Prediction~\citep{zhang2024seer} \\
 & JoinBoost & JoinBoost: Grow Trees over Normalized Data Using Only SQL~\citep{huang2023joinboost} \\
\addlinespace[2pt]
Machine Learning & LeJEPA & When Does LeJEPA Learn a World Model?~\citep{klindt2026lejepa} \\
 & Polar Express & The Polar Express: Optimal Matrix Sign Methods and Their Application to the Muon Algorithm~\citep{amsel2025polarexpress} \\
 & DeTok & Latent Denoising Makes Good Tokenizers~\citep{yang2025detok} \\
 & Representation Fr\'echet Loss & Representation Fr\'echet Loss for Visual Generation~\citep{yang2026rfl} \\
 & Flow-OPD & Flow-OPD: On-Policy Distillation for Flow Matching Models~\citep{fang2026flowopd} \\
 & RAGEN & RAGEN: Understanding Self-Evolution in LLM Agents via Multi-Turn Reinforcement Learning~\citep{wang2025ragen} \\
 & DyT & Transformers without Normalization~\citep{zhu2025dyt} \\
 & Nano World Model & Nano World Models: A Minimalist Implementation of Future Video Prediction~\citep{huang2026nanowm} \\
 & KV Compaction & Fast KV Compaction via Attention Matching~\citep{zweiger2026kvcompaction} \\
\addlinespace[2pt]
MLSys & SVG-EAR & SVG-EAR: Parameter-Free Linear Compensation for Sparse Video Generation via Error-aware Routing~\citep{zhou2026svgear} \\
 & FlashLib K-Means & Flash-KMeans: Fast and Memory-Efficient Exact K-Means~\citep{yang2026flashkmeans} \\
 & FlashLib DBSCAN & FlashLib: Bringing Flash Magic to Classical Machine Learning Operators~\citep{yang2026flashlib} \\
 & FlashLib IVF-PQ & FlashLib: Bringing Flash Magic to Classical Machine Learning Operators~\citep{yang2026flashlib} \\
 & FlashLib KNN & FlashLib: Bringing Flash Magic to Classical Machine Learning Operators~\citep{yang2026flashlib} \\
 & Continuum & Continuum: Efficient and Robust Multi-Turn LLM Agent Scheduling with KV Cache Time-to-Live~\citep{li2025continuum} \\
\addlinespace[2pt]
HPC & Hygra & Practical Parallel Hypergraph Algorithms~\citep{shun2020hygra} \\
 & ParButterfly & Parallel Algorithms for Butterfly Computations~\citep{shi2020parbutterfly} \\
 & Parallel DBSCAN & Theoretically-Efficient and Practical Parallel DBSCAN~\citep{wang2020dbscan} \\
 & RegionsSort & Theoretically-Efficient and Practical Parallel In-Place Radix Sorting~\citep{obeya2019regionssort} \\
\addlinespace[2pt]
Programming Languages & BABBLE & babble: Learning Better Abstractions with E-Graphs and Anti-unification~\citep{cao2023babble} \\
 & egg & egg: Fast and Extensible Equality Saturation~\citep{willsey2021egg} \\
 & Polygon & Polygon: Symbolic Reasoning for SQL using Conflict-Driven Under-Approximation Search~\citep{zhao2025polygon} \\
 & Glenside & Pure Tensor Program Rewriting via Access Patterns (Representation Pearl)~\citep{smith2021glenside} \\
\addlinespace[2pt]
Architecture & Clusterwise SpGEMM & Improving SpGEMM Performance Through Matrix-Reordering and Cluster-wise Computation~\citep{islam2025spgemm} \\
 & KAMI & KAMI: Communication-Avoiding General Matrix Multiplication within a Single GPU~\citep{wang2025kami} \\
 & RUNLTS & RUNLTS: Register-value-aware Predictor Utilizing Nested Large Tables~\citep{koizumi2025runlts} \\
\addlinespace[2pt]
Cryptography & Structured LWE & A Systematic Study of Sparse LWE~\citep{jain2024sparselwe} \\
 & QuarterPIR & Efficient Pre-processing PIR Without Public-Key Cryptography~\citep{ghoshal2024quarterpir} \\
 & CKKS Noise & Accurate and Composable Noise Estimates for CKKS with Application to Exact HE Computation~\citep{bossuat2025ckks} \\
\addlinespace[2pt]
Network & BBQ & BBQ: A Fast and Scalable Integer Priority Queue for Hardware Packet Scheduling~\citep{atre2024bbq} \\
 & Shockwave & Shockwave: Fair and Efficient Cluster Scheduling for Dynamic Adaptation in Machine Learning~\citep{zheng2023shockwave} \\
 & MetaOpt & Finding Adversarial Inputs for Heuristics using Multi-level Optimization~\citep{namyar2024metaopt} \\
\addlinespace[2pt]
Security & Progent & Progent: Securing AI Agents with Privilege Control~\citep{shi2025progent} \\
 & LRR & Repairing Regex Vulnerabilities via Localization-Guided Instructions~\citep{sung2026lrr} \\
\addlinespace[2pt]
Robotics & Diffusion Policy & Diffusion Policy: Visuomotor Policy Learning via Action Diffusion~\citep{chi2023diffusionpolicy} \\
 & MuJoCo Playground & Demonstrating MuJoCo Playground~\citep{zakka2025mujocoplayground} \\
\addlinespace[2pt]
Bioinformatics & Edlib & Edlib: a C/C++ library for fast, exact sequence alignment using edit distance~\citep{sosic2017edlib} \\
\addlinespace[2pt]
Software Engineering & CREST & Heuristics for Scalable Dynamic Test Generation~\citep{burnim2008crest} \\
\addlinespace[2pt]
Theory & Binary Fuse Filter & Binary Fuse Filters: Fast and Smaller Than Xor Filters~\citep{graf2022binaryfuse} \\
\end{longtable}
\endgroup

\section{Task resources}
\label{app:resources}

\Cref{tab:resources} lists what each task gives the agent and the verifier. The agent has five hours on every task. GPU tasks give the agent short-lived GPU runners for its own experiments and score submissions on separate verifier GPUs (every A100 is the 80\,GB part); the listed GPU time is a budget for the whole run, not a measured cost. Progent is the only task that also calls a model from inside the task: the agents it secures run on a task model with their own per-run spending limit.

\begingroup\footnotesize\setlength{\LTcapwidth}{\linewidth}\setlength{\LTleft}{0pt}\setlength{\LTright}{0pt}\setlength{\tabcolsep}{3pt}
\captionof{table}{\textbf{Task resources.} What each task gives the agent and the verifier. Agent cores and memory: the container the agent works in, memory in GiB. GPU: the GPU type of the task; a dash means the task uses no GPU. GPU-hours: the total the agent may use for its own experiments, and the total for scoring its submissions; $n\times h$ means $n$ GPUs at a time for $h$ hours each. Train evals: the most submissions the agent may score on the development workloads. Final eval time limit: how long the hidden final evaluation may run, in minutes.}\label{tab:resources}\nopagebreak
\begin{longtable}{@{\extracolsep{\fill}}llrrcrrrr@{}}
\toprule
 & & \multicolumn{2}{c}{Agent} & & \multicolumn{2}{c}{GPU-hours} & Train & Final eval \\ \cmidrule(lr){3-4}\cmidrule(lr){6-7}Domain & Task & Cores & Memory & GPU & Agent & Verifier & evals & time limit \\
\midrule
\endfirsthead
\toprule
 & & \multicolumn{2}{c}{Agent} & & \multicolumn{2}{c}{GPU-hours} & Train & Final eval \\ \cmidrule(lr){3-4}\cmidrule(lr){6-7}Domain & Task & Cores & Memory & GPU & Agent & Verifier & evals & time limit \\
\midrule
\endhead
\bottomrule
\endfoot
Database & ScaleGPM & 8 & 16 & -- & -- & -- & 100 & 15 \\
 & Yannakakis & 2 & 8 & -- & -- & -- & 100 & 15 \\
 & FARGO Query Stopping & 4 & 4 & -- & -- & -- & 100 & 15 \\
 & Spectrum & 4 & 8 & -- & -- & -- & 100 & 60 \\
 & SCOPE & 1 & 4 & -- & -- & -- & 100 & 30 \\
 & Persistent Treap & 8 & 16 & -- & -- & -- & 100 & 30 \\
 & FARGO Query Core & 4 & 4 & -- & -- & -- & 50 & 15 \\
 & Seer & 8 & 4 & -- & -- & -- & 50 & 30 \\
 & JoinBoost & 2 & 16 & -- & -- & -- & 100 & 120 \\
Machine Learning & LeJEPA & 2 & 12 & -- & -- & -- & 8 & 6.7 \\
 & Polar Express & 2 & 4 & -- & -- & -- & 100 & 30 \\
 & DeTok & 3 & 12 & L4 & 1.5 & 1.5 & 6 & 90 \\
 & Representation Fr\'echet Loss & 3 & 12 & A100 & 2 & 9 & 3 & 180 \\
 & Flow-OPD & 3 & 12 & H100 & 0.83 & 4.17 & 4 & 50 \\
 & RAGEN & 3 & 12 & A100 & 0.5 & 2 & 3 & 30 \\
 & DyT & 3 & 12 & L4 & 0.5 & 1 & 3 & 15 \\
 & Nano World Model & 3 & 12 & A10G & 0.5 & 5.67 & 3 & 170 \\
 & KV Compaction & 4 & 16 & H100 & 5 & 8$\times$3 & 8 & 90 \\
MLSys & SVG-EAR & 3 & 12 & H100 & 15 & 4$\times$91 & 25 & 210 \\
 & FlashLib K-Means & 4 & 8 & H100 & 3 & 5 & 25 & 30 \\
 & FlashLib DBSCAN & 4 & 8 & H100 & 3 & 3.67 & 25 & 40 \\
 & FlashLib IVF-PQ & 4 & 8 & H100 & 3 & 7 & 25 & 240 \\
 & FlashLib KNN & 4 & 8 & H100 & 3 & 4.5 & 25 & 90 \\
 & Continuum & 8 & 32 & H100 & 5 & 4$\times$48 & 25 & 1800 \\
HPC & Hygra & 8 & 4 & -- & -- & -- & 100 & 30 \\
 & ParButterfly & 4 & 8 & -- & -- & -- & 100 & 30 \\
 & Parallel DBSCAN & 8 & 4 & -- & -- & -- & 100 & 10 \\
 & RegionsSort & 4 & 2 & -- & -- & -- & 100 & 15 \\
Programming Languages & BABBLE & 2 & 8 & -- & -- & -- & 100 & 10 \\
 & egg & 8 & 4 & -- & -- & -- & 100 & 30 \\
 & Polygon & 2 & 8 & -- & -- & -- & 100 & 5.5 \\
 & Glenside & 2 & 4 & -- & -- & -- & 100 & 5 \\
Architecture & Clusterwise SpGEMM & 2 & 16 & -- & -- & -- & 3 & 120 \\
 & KAMI & 3 & 4 & H100 & 0.25 & 0.5 & 12 & 30 \\
 & RUNLTS & 1 & 4 & -- & -- & -- & 100 & 120 \\
Cryptography & Structured LWE & 8 & 32 & -- & -- & -- & 100 & 2 \\
 & QuarterPIR & 4 & 8 & -- & -- & -- & 8 & 30 \\
 & CKKS Noise & 2 & 2 & -- & -- & -- & 100 & 10 \\
Network & BBQ & 2 & 4 & -- & -- & -- & 100 & 30 \\
 & Shockwave & 4 & 4 & -- & -- & -- & 3 & 30 \\
 & MetaOpt & 4 & 8 & -- & -- & -- & 100 & 60 \\
Security & Progent & 4 & 8 & -- & -- & -- & 5 & 180 \\
 & LRR & 4 & 12 & L4 & -- & 2 & 100 & 60 \\
Robotics & Diffusion Policy & 2 & 4 & -- & -- & -- & 100 & 62 \\
 & MuJoCo Playground & 3 & 12 & H100 & 4 & 8.5 & 3 & 240 \\
Bioinformatics & Edlib & 2 & 4 & -- & -- & -- & 20 & 10 \\
Software Engineering & CREST & 8 & 16 & -- & -- & -- & 100 & 180 \\
Theory & Binary Fuse Filter & 2 & 4 & -- & -- & -- & 100 & 3 \\
\end{longtable}
\addtocounter{table}{-1}
\endgroup

\section{Agent settings}
\label{app:agents}

\Cref{tab:agents} lists each agent's harness version, reasoning effort and output token limit in the evaluation. The effort and the limit are the values the harness sent with its model requests; for Claude Code, the effort is the one we set when launching it. An empty cell marks a value we could not confirm.

Some harnesses change a setting from call to call. Muse Code adds short side calls that watch the main agent and decide whether to remind it of something; these run at low or medium effort, and the table gives the effort of the main agent's calls. Pi sets Gemini 4 Argon's output limit to the room left in the context window. Kimi Code gives Kimi K2.7 the smaller of the listed limit and the room left. ZCode and Step Code also shrink their limit to the room left, which happens only when the context window is nearly full.

\begin{table}[t]
\caption{\textbf{Agent settings.} Harness and version of each agent, with the reasoning effort and output token limit the harness requested. ``Not set'' leaves the default to the provider. Rows follow FCI order (\Cref{sec:exp-eci}); Kimi K2.7 serves only as the 0 point of FCI.}
\label{tab:agents}
\centering\small
\begin{tabular}{@{}llllr@{}}
\toprule
Model & Harness & Version & Reasoning effort & Output limit \\
\midrule
Claude Opus 5.5 & Claude Code & 2.1.296 & max &  \\
GPT-6 Astra & Codex CLI & 0.162.1 & xhigh & not set \\
GPT-6.1 Sol & Codex CLI & 0.162.1 & xhigh & not set \\
Gemini 4 Argon & Pi & 1.2.0 & high & rest of context \\
GLM 5.3 & ZCode & 0.16.9 & max & 262,144 \\
Muse Wonder 1.0 & Muse Code & 1.4.2 & max & 128,000 \\
DeepSeek V4.1 Flash & DeepSeek Harness & 0.2.0-rc.2 & max & 262,144 \\
Kimi K3 & Kimi Code & 0.31.0 & high & 131,072 \\
Qwen 3.8 Max & Qwen Code & 0.25.0 & high & not set \\
Muse Spark 1.3 & Muse Code & 1.4.2 & high & 128,000 \\
GLM 5.3 Flash & ZCode & 0.16.9 & max & 262,144 \\
Step 5 Preview & Step Code & 0.1.0 & high & 64,000 \\
Kimi K2.7 & Kimi Code & 0.31.0 & medium & up to 131,072 \\
\bottomrule
\end{tabular}
\end{table}

\section{Computing FCI and FDI}
\label{app:fci}

FCI and FDI come from one statistical model, fitted to all runs at once. Our fitting code carries out the twelve steps below, and every value in this appendix comes from the fit behind \Cref{sec:exp-eci}. \Cref{tab:fci-params} lists the constants and fitted values.

\paragraph{Step 1: collect the runs.} The fit uses every run with a final score on the evaluated tasks. Besides the 12 reported agents, it includes Kimi K2.7, which anchors the zero of the scale but is not reported. A run that ends without a saved final submission scores 0, and so do the two runs that Gemini 4 Argon's provider refused (\Cref{sec:exp-setup}). This gives 624 runs.

\paragraph{Step 2: score each metric.} Each task's judge measures a fixed set of metrics, such as latency, throughput or accuracy, on every held-out workload. A run beats a metric when its final submission beats the human reference on that metric in every held-out workload. A run that fails before the held-out evaluation beats no metric. A run therefore passes exactly when it beats every metric of its task.

\paragraph{Step 3: merge metrics into groups.} Two metrics that every run of a task beats or misses together cannot be told apart by the data, and fitting each with its own difficulty would make the task look harder than it is. Such metrics form one group. A run beats a group when it beats every member. Let $K_t$ be the number of groups of task $t$, and let $x_{atg} \in \{0, 1\}$ record whether agent $a$ beats group $g$ of task $t$. The tasks yield 115 groups.

\paragraph{Step 4: measure the score margin.} Final scores use each task's own scale, so the fit converts a run's distance from the human reference into a common unit. The reference score $s^{\mathrm{ref}}_t$ is the median, over the task's runs, of the human reference's final score measured within each run. If the runs of a task record no positive reference score, a value set in advance takes its place. The spread $d_t$ is the standard deviation of the task's final scores that lie strictly between 0 and 100, or of all its final scores when fewer than three do. The unit is $c_t = \max(d_t, c)$, whose floor $c$ is the median of $d_t$ over the tasks. The margin of agent $a$ on task $t$ is
\begin{equation}
y_{at} = \frac{s_{at} - s^{\mathrm{ref}}_t}{c_t},
\label{eq:fci-margin}
\end{equation}
where $s_{at}$ is the run's final score. A run enters the fit with its margin in one of three ways. A run that passes, or fails and scores below the reference, contributes $y_{at}$ as measured. A run that fails but scores above the reference contributes only $y_{at} \geq 0$: it is at least level with the reference, since at least one of its metric groups misses. A run that ties the reference exactly contributes no margin.

\paragraph{Step 5: write the likelihood of one run.} Each agent has a strength $\theta_a$, each metric group a difficulty $\beta_{tg}$, and each task a tie level $h_t$, all in logits. The tie level is the strength at which a run's expected margin is zero. A run also has a run effect $u_{at} \sim \mathcal{N}(0, \sigma_u^2)$, which says how far this run lands above or below its agent's strength. The groups and the margin come from the same measurements, so both share the run effect. Given $u_{at} = u$, the run beats group $g$ with probability
\begin{equation}
P(x_{atg} = 1 \mid u) = \sigma(\theta_a + u - \beta_{tg}), \qquad \sigma(z) = \frac{1}{1 + e^{-z}},
\label{eq:fci-group}
\end{equation}
and its margin is normal with slope $\alpha$ and noise $\sigma_e$,
\begin{equation}
y_{at} \mid u \sim \mathcal{N}\big(\alpha\,(\theta_a + u - h_t),\ \sigma_e^2\big).
\label{eq:fci-y}
\end{equation}
A measured margin enters through this density, and a margin known only to be at least 0 through the probability $\Phi\big(\alpha\,(\theta_a + u - h_t) / \sigma_e\big)$, where $\Phi$ is the standard normal distribution function. Write $f_{at}(u)$ for this density, this probability, or 1 for a run that ties. The likelihood of the run integrates the run effect out,
\begin{equation}
L_{at} = \int \prod_{g=1}^{K_t} \sigma(\theta_a + u - \beta_{tg})^{x_{atg}} \big(1 - \sigma(\theta_a + u - \beta_{tg})\big)^{1 - x_{atg}} \, f_{at}(u) \, \phi_{\sigma_u}(u) \, du,
\label{eq:fci-run}
\end{equation}
where $\phi_{\sigma_u}$ is the density of $\mathcal{N}(0, \sigma_u^2)$. Gauss--Hermite quadrature with 64 nodes evaluates the integral.

\paragraph{Step 6: set the priors.} Each strength has the prior $\theta_a \sim \mathcal{N}(0, 1/\lambda)$ with $\lambda = 0.35$. Each task's tie level has $h_t \sim \mathcal{N}(0, 6^2)$, and its group difficulties have $\beta_{tg} \sim \mathcal{N}(h_t + \delta, \tau^2)$. The offset $\delta$ is how far a typical group sits above the tie level, and $\tau$ is how widely the groups scatter around that point. This prior keeps a group that no run beats at a finite difficulty, pulled toward the task's other evidence.

\paragraph{Step 7: find the joint mode.} For fixed hyperparameters $(\sigma_u, \alpha, \delta, \tau, \sigma_e)$, the fit maximizes the log posterior
\begin{equation}
\sum_{a, t} \log L_{at} + \sum_a \log p(\theta_a) + \sum_t \log p(h_t, \beta_{t1}, \ldots, \beta_{tK_t})
\label{eq:fci-posterior}
\end{equation}
jointly over the 13 strengths and the 163 task parameters, namely 115 group difficulties and 48 tie levels. It uses Newton's method. Each run touches one strength and one task, so the Hessian is sparse: the step for the strengths is solved through their Schur complement, and a step that lowers the objective is halved until it does not.

\paragraph{Step 8: fit the hyperparameters.} The five hyperparameters maximize an approximate marginal likelihood. At the joint mode of Step 7, a Laplace approximation integrates each task's parameters out; the sum of these terms over tasks, plus the strengths' log prior at the mode, is the objective. Nelder--Mead searches it from two starting points, with $\sigma_u$ capped at 7 logits. \Cref{tab:fci-params} lists the values at its maximum, and Step 7 is then repeated at these values.

\paragraph{Step 9: place the agents on the FCI scale.} An agent's FCI is its strength relative to Kimi K2.7's,
\begin{equation}
\mathrm{FCI}_a = 10\,(\theta_a - \theta_{\mathrm{K2.7}}),
\label{eq:fci-scale}
\end{equation}
where $\theta_{\mathrm{K2.7}}$ is the strength of Kimi K2.7. Kimi K2.7 thus sits at 0, and ten points equal one logit: an agent ten points higher has $e \approx 2.72$ times the odds of beating any metric group, for the same run effect.

\paragraph{Step 10: bootstrap the FCI intervals.} Each of 500 bootstrap draws resamples the tasks with replacement, so a task may appear several times or not at all. The draw refits the strengths and task parameters from the full fit's mode, with the hyperparameters held at their Step 8 values, and maps its strengths to FCI through its own $\theta_{\mathrm{K2.7}}$. An agent's 90\% interval spans the 5th to 95th percentile of its FCI across the draws. All 500 draws converge.

\paragraph{Step 11: compute each task's FDI.} With the group difficulties at the mode, an agent of strength $\theta$ passes task $t$ with probability
\begin{equation}
P_t(\theta) = \int \prod_{g=1}^{K_t} \sigma(\theta + u - \beta_{tg}) \, \phi_{\sigma_u}(u) \, du,
\label{eq:fci-pass}
\end{equation}
since a pass needs every group beaten in the same run. $P_t$ rises with $\theta$, and the task's difficulty $D_t$ solves $P_t(D_t) = 1/2$, found by bisection. Its FDI is $10\,(D_t - \theta_{\mathrm{K2.7}})$, on the scale of \Cref{eq:fci-scale}. For the 90\% interval of the FDI, 2,000 draws of the task's parameters come from the Laplace Gaussian at the mode, with the strengths held. Each draw is weighted by the ratio of the exact posterior to the Gaussian density, which corrects the one-sided posterior of a group that no run beats. The interval spans the weighted 5th to 95th percentiles of $D_t$ across the draws.

\paragraph{Step 12: refit FCI under a cost budget.} Test-time scaling, defined in \Cref{sec:exp-rules}, holds every task's parameters and the hyperparameters at the full fit. At a budget of $B$ US\$ per run, a run that finished within $B$ contributes its outcomes from Steps 2 to 4. A run still going at $B$ counts as a run with no final submission: it beats no group, and its margin is that of a final score of 0. The agent's FCI at $B$ is the strength that maximizes its runs' log likelihood plus its log prior, found by Newton's method and mapped by \Cref{eq:fci-scale}. Its 90\% interval comes from 100 bootstrap draws over the agent's tasks. The rule uses no development score, because development scores swing: the judge assigns 0 whenever a submission fails a correctness or quality check, yet most runs that drop to 0 after a positive score recover above 0 later. A rule based on development scores would also have to pick a patch at each budget (the newest, the best so far or the last valid one) without held-out verification.

\begin{table}[t]
\caption{\textbf{Constants and fitted values of the FCI model.} Fixed constants are set before the fit; fitted values come from Step 8 or, for the unit floor $c$, from the data in Step 4. Values in logits are also given in FCI points, at ten points per logit.}
\label{tab:fci-params}
\centering\small
\begin{tabular}{@{}llrr@{}}
\toprule
Symbol & Meaning & Logits & FCI points \\
\midrule
\multicolumn{4}{@{}l}{\emph{Fixed}} \\
$\lambda$ & Precision of the strengths' prior & 0.35 & \\
 & Prior standard deviation of a tie level $h_t$ & 6 & 60 \\
 & Gauss--Hermite nodes & 64 & \\
 & Bootstrap draws for FCI, and for FCI under a budget & 500, 100 & \\
 & Weighted draws for an FDI interval & 2,000 & \\
\midrule
\multicolumn{4}{@{}l}{\emph{Fitted}} \\
$\sigma_u$ & Standard deviation of the run effect & 2.041 & 20.41 \\
$\alpha$ & Slope of the margin per logit & 0.597 & \\
$\delta$ & Offset of a typical group above the tie level & 1.276 & 12.76 \\
$\tau$ & Spread of group difficulties & 3.954 & 39.54 \\
$\sigma_e$ & Noise of the margin, in margin units & 2.323 & \\
$c$ & Floor of the margin unit, in score points & 5.78 & \\
\bottomrule
\end{tabular}
\end{table}

\end{document}